\documentclass[10pt,twocolumn,letterpaper]{article}

\usepackage[pagenumbers]{cvpr} 
\usepackage[accsupp]{axessibility}  
\usepackage{graphicx}
\usepackage{pifont}
\usepackage{amsmath}
\usepackage{amssymb}
\usepackage{ tipa }
\usepackage{booktabs}
\usepackage{tabularx}
\usepackage{multicol}
\usepackage{multirow}
\usepackage[dvipsnames]{xcolor}

\usepackage[pagebackref,breaklinks,colorlinks]{hyperref}
\usepackage{colortbl}

\usepackage[capitalize]{cleveref}
\crefname{section}{Sec.}{Secs.}
\Crefname{section}{Section}{Sections}
\Crefname{table}{Table}{Tables}
\crefname{table}{Tab.}{Tabs.}

\usepackage{color}
\newcommand{\YSR}[1]{\textcolor{blue}{YSR: #1}}

\newcommand{\cmark}{\ding{51}}%
\newcommand{\xmark}{\ding{55}}%
\def\cvprPaperID{*****} 
\def\confName{CVPR}
\def\confYear{2022}

\begin{document}

\title{
Video Action Detection: Analysing Limitations and Challenges
}
\author{Rajat Modi\textsuperscript{$\dagger$} , Aayush Jung Rana\textsuperscript{$\dagger$}, Akash Kumar\textsuperscript{$\dagger$}, Praveen Tirupattur\textsuperscript{$\dagger$},\\
    Shruti Vyas\textsuperscript{$\ddagger$}, Yogesh Singh Rawat\textsuperscript{$\ddagger$} \& Mubarak Shah\textsuperscript{$\ddagger$}\\
Center for Research in Computer Vision\\
University of Central Florida, Orlando, Florida, USA\\
{\tt\small {\textsuperscript{$\dagger$}\{rajatmodi, aayushjr, akash\_k, praveentirupattur\}}@knights.ucf.edu}\\
{\tt\small {\textsuperscript{$\ddagger$}\{shruti, yogesh, shah\}}@crcv.ucf.edu}
}
\maketitle

\begin{abstract}
Beyond possessing large enough size to feed data hungry machines (eg, transformers), what attributes measure the quality of a dataset? Assuming that the definitions of such attributes do exist, how do we quantify among their relative existences? Our work attempts to explore these questions for video action detection. The task aims to spatio-temporally localize an actor and assign a relevant action class. 
We first analyze the existing datasets on video action detection and discuss their limitations. Next, we propose a new dataset, Multi Actor Multi Action (MAMA) which overcomes these limitations and is more suitable for real world applications.
In addition, we perform a biasness study which analyzes a key property differentiating videos from static images: the temporal aspect. This reveals if the actions in these datasets really need the motion information of an actor, or whether they predict the occurrence of an action even by looking at a single frame. Finally, we investigate the widely held assumptions on the importance of temporal ordering: is temporal ordering important for detecting these actions? Such extreme experiments show existence of biases which have managed to creep into existing methods inspite of careful modeling. 
The dataset and code is publicly available for research at \footnote{https://www.crcv.ucf.edu/research/projects/mama-multi-actor-multi-action-dataset-for-action-detection/} 

\end{abstract}

\section{Introduction}
\label{sec:intro}

\begin{figure}[t]
  \centering
\includegraphics[width=1\linewidth]{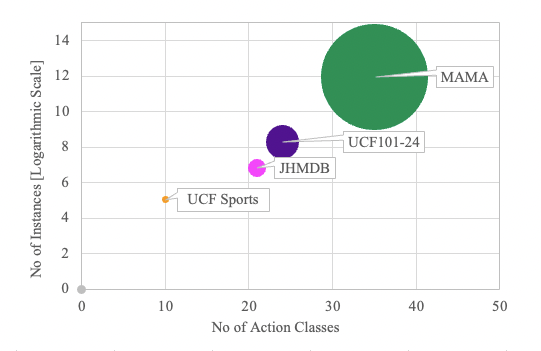}

   \caption{An illustration of relative properties of datasets in activity detection. Size of the bubble corresponds to the no of samples in the dataset. X axis: No of action classes. MAMA has more classes [35] than standard detection datasets. Y axis: No of Instances. MAMA has a lot more actors, including crowded scenarios. Finally, MAMA possesses 10x more samples. 
   }
   \label{fig:bubbledia}
\end{figure}

\begin{figure*}
    \centering
    \includegraphics[width=\linewidth]{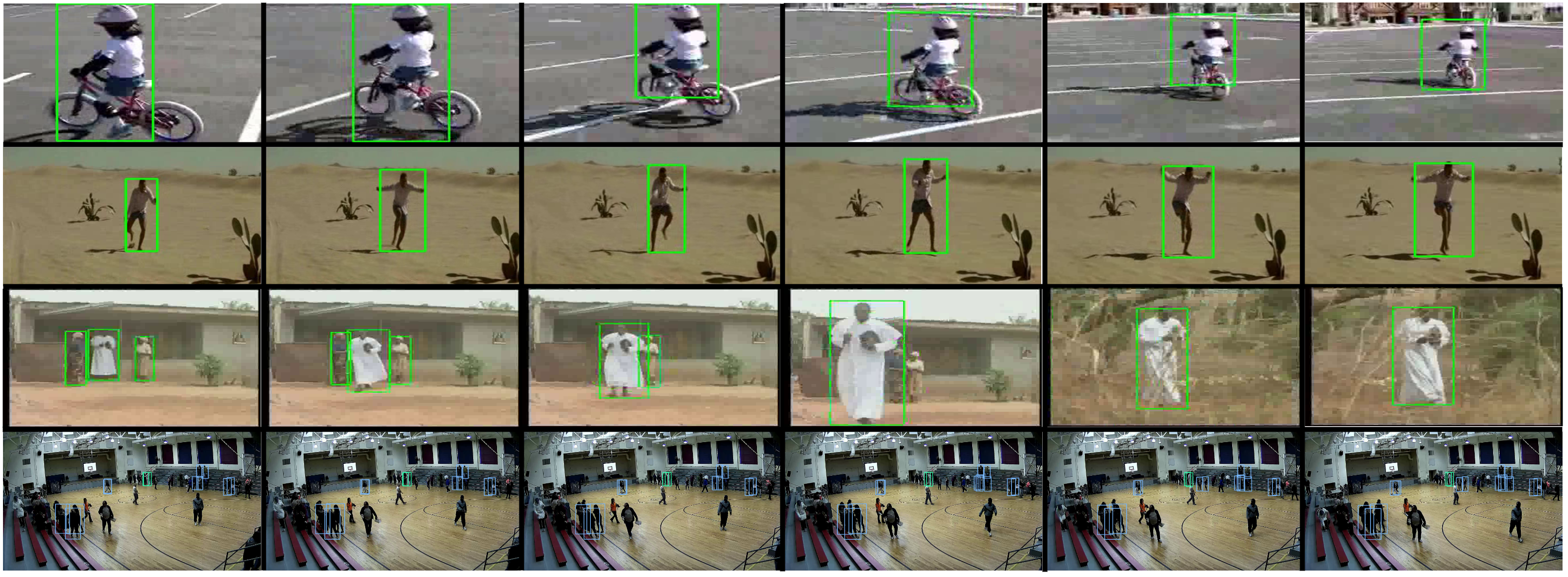}
    \caption{Each row shows samples from UCF101, JHMDB, AVA, and MAMA respectively. UCF101 (First Row): contains a single activity instances per video. Most samples show an actor in far closer perspective than observed in real scenarios. JHMDB (Second Row): has even lesser videos than UCF101. AVA (Third Row) Since most of the frames are annotated at 1 sec intervals, successive annotated frames have significant differences in the actor motion.  MAMA(Fourth Row): Shows even far off actors being annotated, in a crowded setting. MAMA can possesses multiple activities per video. 
    }
    \label{fig:samples*}
\end{figure*}

Video action understanding has been widely explored in recent years in terms of action classification \cite{carreira2017quo_i3d,tran2018closer_r2p1d,hara2018can_r3d,feichtenhofer2019slowfast,wang2018nonlocal,xie2018rethinking,soomro2014sports,tran2015c3dfb,vyas2020multi} and detection \cite{yang2019step,duarte2018videocapsulenet,hou2017tcnn,weinzaepfel2015learning,li2018recurrent,kalogeiton2017action,rana2021we,el2018real,kumar2022end,kumar2022end} for a wide range of actions. Analyzing the videos for action classification task involves feature extraction and prediction on a video-level, with recent methods achieving state-of-the-art performance \cite{tran2015c3dfb,carreira2017quo_i3d,feichtenhofer2019slowfast}. 
Compared to action classification, action detection is a harder task as it requires the spatio-temporal prediction along with the action class prediction for \textit{each} action in the video. The increased complexity in action detection has garnered more interest recently with significant progress in various datasets \cite{yang2019step,hou2017tcnn,kalogeiton2017action,rana2021we,weinzaepfel2016human,mcintosh2020visual}. 
Spatio-temporal detection for action requires datasets with frame-level annotations (bounding-box, pixel-wise) which is costly to produce, resulting in limited datasets that can be used for this task. Current datasets annotated for video action detection include dense frames annotation (UCF101-24 \cite{soomro2012ucf101}, JHMDB-21 \cite{jhuang2013towards_jhmdb}, UCF-Sports \cite{rodriguez2008action_ucfsports}, VIRAT \cite{oh2011large_virat}, MEVA \cite{corona2021meva}) and sparse frames annotation (A2D \cite{xu2015can}, DALY \cite{weinzaepfel2016human}, AVA \cite{gu2018ava}), which is limited compared to the vast action classification datasets due to higher spatio-temporal annotation costs. A  graphical comparison of these datasets has been illustrated in Figure \ref{fig:bubbledia}. 


Datasets with dense spatio-temporal annotations have bounding-box or pixel-wise annotation for each action/actor for the entire duration of the action (UCF101-24, JHMDB-21, VIRAT, MEVA), with most datasets being untrimmed. While this gives better results on supervised methods, it is costly to annotate all the frames. UCF101-24 and JHMDB-21 are the two widely used dense annotation datasets for action detection and have saturated with recent methods performing exceedingly well \cite{duarte2018videocapsulenet,yang2019step}. VIRAT and MEVA have larger untrimmed dense annotations from surveillance videos which is significantly more costly to produce compared to UCF101-24, but have low adoption in research community since they make action detection harder due to smaller actors, simultaneous multiple actions and untrimmed long videos \cite{yu2022argus++,rizve2021gabriella,dave2022gabriellav2}.
Sparse annotation datasets (A2D, DALY, AVA) reduces annotation cost and allows annotating more samples, however they are harder to train for action detection task with current methods due to not having sufficient temporal annotation for doing action detection \cite{yang2019step,feichtenhofer2019slowfast,weinzaepfel2016human,fan2021multiscale,rana2021we}. 
We analyze the properties of these various datasets and bring to light their similarities and differences in order to understand how action detection can be improved as a whole. Each dataset provides different context for the actions in them (eg. movie scenes, surveillance, controlled lab, YouTube videos, sports clips), with benefits and limitations for action detection generalization. 


To better understand the importance of temporal annotation along with spatial annotation for better action detection, we analyze the effect temporal information has in different datasets and evaluate properties such as temporal aspect and order. It is observed that temporal information has positive correlation for action classification \cite{carreira2017quo_i3d,hara2018can_r3d}. We look into what type of relations temporal information has for action detection task and provide insight into how each dataset contributes to this task. 
To further improve action detection in real-world scenarios, we introduce the \textbf{MAMA} dataset. We aim to provide more real world scenarios with multiple simultaneous actions, mitigating shortcomings of prior detection datasets. In summary, we contribute via \textsc{(1)} a new dataset addressing issues in existing datasets; \textsc{(2)} study of temporal aspect and order for action detection in all datasets; and \textsc{(3)} in depth analysis in similarities/differences of all video action datasets.



\section{Video action detection datasets}

\begin{table*}[]

\resizebox{\textwidth}{!}{\begin{tabular}{l|>{\color[gray]{0.7}}c|>{\color[gray]{0.7}}c|>{\color[gray]{0.7}}c|c|c|c|c}
\textbf{Datasets} & \textbf{VIRAT} & \textbf{MEVA} & \textbf{AVA} & \textbf{UCF Sports} & \textbf{UCF101-24} & \textbf{JHMDB}  & \textbf{MAMA} \\
\hline
\#Classes & 40 & 39 & 80 & 10 & 24 & 21 &  35 \\
Source & Surveillance & Surveillance & Movies  & Sports   & Sports  & Movies  & Surveillance \\
Resolution  & 1920 $\times$ 1080  & 1920 $\times$ 1080  & 320 $\times$ 400  & 690 $\times$ 450 & 320 $\times$ 240 & 320 $\times$ 240  &  1920 $\times$ 1080  \\
Total Videos & 369 & 1145 & 430 & 150  & 3194 & 928 & 32726 \\
Total Frames & 1.05M & 10.31M & $\sim$ 11M & 10K & 558K & 32K & 2.7M \\
Avg. Video Length & 1.6 min & 5 min & 15min & 5.8 sec  & 5.8 sec & 1.4 sec & 2.6 sec  \\
Avg. Action Duration & 10.1 sec & 10.3 sec & - & 5.8sec   & 4.5 sec  &1.4 sec  & 2.6 sec \\
Total \#Instances & 8K & 37K & 1.62M & 154 & 4030 & 928  &  32726  \\
Multi-Actor & \cmark & \cmark  & \cmark & \cmark  & \cmark  & \cmark  & \cmark \\ 
Multi-Label & \cmark & \cmark  & \cmark & \xmark  & \xmark  & \xmark  & \cmark \\  
Annotation Type &  Boxes &   Boxes  & Boxes &  Boxes &  Boxes & Pixels  &  Boxes\\ 
Spatio-Temporal Ann. & \cmark  &  \cmark   & \xmark & \cmark  & \cmark  & \cmark  & \cmark \\
Class Distribution & Long-tail  &  Long-tail   & Long-tail & Uniform  & Uniform  &  Uniform &  Long-tail\\
\end{tabular}}
\caption{ A relative comparison of datasets for brevity.  Three columns from the left are grey because although they are standard datasets in action-detection, they don't meet the criteria for this biasness study. VIRAT/MEVA contain untrimmed videos. AVA only annotates frames at 1 sec intervals, therefore it is not truly spatio-temporal in nature. Notice how MAMA exceeds in several statistics to all the other datasets.
}
\label{tab:comparison}
\end{table*}

Initial research on video action understanding focused on action classification where the task is to identify the action in a short, manually trimmed video containing a single action. Some of the popular action classification datasets are HMDB \cite{kuehne2011hmdb}, UCF101 \cite{soomro2012ucf101}, Sports-1M \cite{karpathy2014large}, Moments in Time \cite{monfort2019moments}, TinyVIRAT \cite{demir2021tinyvirat,tirupattur2021tinyaction} and Kinetics \cite{carreira2017quo_i3d}. Another action understanding problem that received interest is the temporal action localization, where the task is to detect the temporal extents of actions in a long untrimmed video. ActivityNet\cite{caba2015activitynet}, THUMOS\cite{idrees2017thumos}, MultiTHUMOS \cite{yeung2018every} and Charades\cite{sigurdsson2016hollywood}  are some of the popular temporal action localization datasets. Compared to action classification and temporal action localization, action detection is a harder task and it involves finding both the spatial and temporal extents of actions in untrimmed videos. Some of the datasets that provide spatio-temporal annotations required to address this problem are UCF Sports \cite{rodriguez2008action_ucfsports}, JHMDB \cite{jhuang2013towards_jhmdb}, UCF101-24 \cite{soomro2012ucf101}, AVA \cite{gu2018ava}, VIRAT \cite{oh2011large_virat} and MEVA \cite{corona2021meva}. 

UCF Sports consists of 150 videos from 10 action classes and JHMDB contains a total of 928 videos with 21 action classes. UCF101-24 consists of 3207 videos with annotations for 24 action classes. These datasets provide annotations for each frame, but contain smaller number of actions, fewer number of video and with videos of shorter duration. The widely AVA dataset contains 430 videos, each video of length 15 minutes, and provides annotations for 80 atomic actions. In this dataset, the annotations are not provided for each frame but for a single frame at one seco   nd intervals. VIRAT provides annotations for 40 actions and contains 118 videos in the train/validation set and 246 videos in held-out test set. All the samples in this dataset are of long untrimmed videos of varying length. MEVA consists of 1056 videos, each 5 minutes long, with annotations for 37 actions. Videos in this dataset cover both indoor and outdoor scenes. In Figure \ref{fig:samples*} we show a sample frame from each of the action detection datasets to highlight the differences.

\subsection{Limitations}
Commonly considered sources for building action detection datasets are videos of sport activities, movie scenes or surveillance videos. Compared to videos of sport activities, action detection in videos from movies is much more challenging due to variations in background, view point, scale, and occlusion. However, most of the videos from the movies have a narrow field-of-view and the focus is on the actor/actors performing the action. Surveillance videos on the other hand, have a wider field-of-view and the actions in these videos can occur at different spatial locations. Apart from the source, these datasets also vary in the type of actions, number of actors and the granularity of annotations. The datasets UCF-Sports, UCF101-24 contain composite actions (e.g., pole vaulting) with single actor and bounding box annotations for each frame; AVA is focused on atomic actions (e.g., stand) with multiple actors and provide bounding box annotations for all the actors in a single frame at every one second intervals; JHMDB contains both atomic and composite actions with single actor and pixel-level annotations for every frame. VIRAT and MEVA have both atomic and composite actions with multiple actors and provide annotations for every actor in each frame. Another aspect in which these datasets vary is the distribution of samples. While the small scale datasets (UCF Sports, UCF101-24 and JHMDB) contain similar number of samples for each class, the large-scale datasets (AVA, VIRAT, MEVA) have a long-tail distribution. Please refer to Table \ref{tab:comparison}, for a detailed comparison of the datasets.

\begin{figure*}
    \centering
    \includegraphics[width=0.9\linewidth]{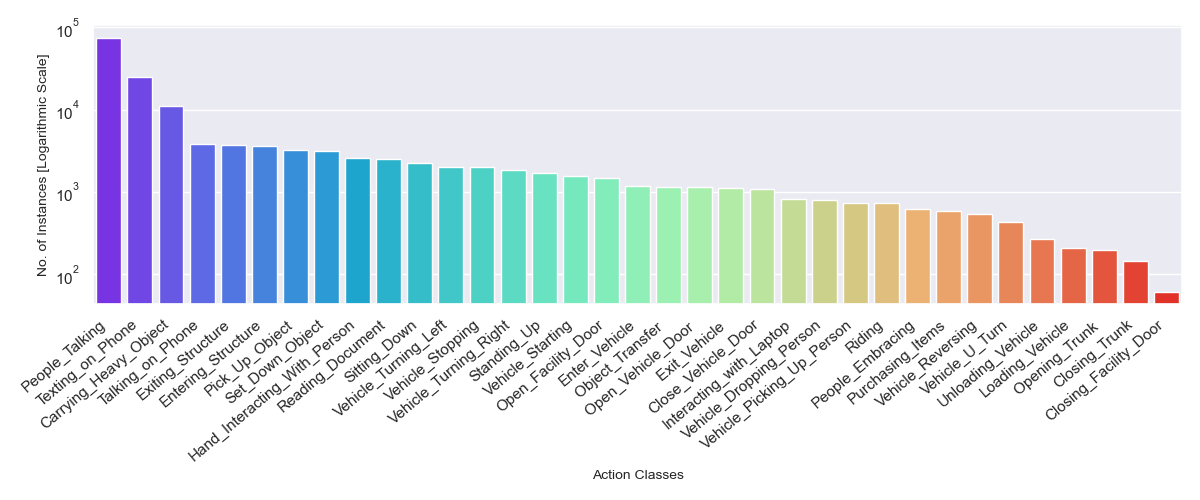}
    \caption{A Bar plot showing the relative instance counts of each of the activities in the MAMA dataset. The bar heights (\# instances) are shown in log scale.}
    \label{fig:activities}
\end{figure*}
\section{MAMA dataset}
Most of the footage relevant for action detection comes from CCTV cameras typically mounted at higher altitudes. Naturally, the videos obtained are untrimmed which can only be processed in chunks by existing action detection models . However, the  activities in such chunks are generally sparse, i.e. in most of the temporally sampled clips of smaller durations, no activity occurs. [eg, very few activities occurs in the night]. Therefore, we propose MAMA dataset which generates shorter temporal crops from untrimmed footage which are more realistic for modelling, and yet sufficiently challenging.

To build the intuition for our dataset, we first explain the temporal trimming protocol. Then, we explore the annotation properties of our dataset, along with the relative comparison of difficulty of MAMA dataset over existing action detection datasets. Finally, we discuss the evaluation protocol on our dataset which relies on standard detection based metrics.


\subsection{Temporal Trimming Protocol}

We model the problem of generating trimmed clips as a variant of the classic \textit{scheduling problem}. Given original untrimmed videos from VIRAT/MEVA dataset, we first generate the [start,end] intervals for each of the activities. A min heap is used to store the activities in ascending order of finish times. For each activity interval, activities lasting lesser than 30 frames are dropped. If an activity lasts longer than 150 frames, we randomly generate a temporal crop between [30,150] frames. Finally, we utilize FFMPEG to actually generate clips from untrimmed video based on given temporal slices. 

This simple, yet effective protocol guarantees that we make a temporal crop, whenever an activity is just starting. The lower constraint of 30 frame generates clips of ideal length so that enough temporal changes could be captured by a model. Varying lengths of clips in our dataset capture both the atomic actions like \textit{Opening a Car Trunk} and long term actions like \textit{Texting on a Cell Phone}. We believe that this should motivate the development of techniques which attend to different length of temporal slices simultanoeusly  based on whether an action is atomic/long term.

To maintain the fairness of our dataset, i.e. the participants don't synthetically \textit{rejoin} temporal crops of a same video to learn more long term context, we anonymize and shuffle the generated clips using a rotating hex cipher. Finally, we assign an 80/20 split of our generated clips using a weighed sampler for extreme multi labelled data. We ensure that the clips belonging to a same video footage go to only one of the splits.

\subsection{Dataset Description}

\cref{fig:activities} shows the relative ratios of 35 classes in the MAMA dataset on a logarithmic scale. The most common activity in the MAMA dataset is \textit{People Talking}, with the least common being \textit{Closing Facility Door}. One interesting aspect is that most of the activities lasting for longer time durations are the ones which are more frequently observed. The activities which are instantaneous (eg \textit{closing facility door}) are concentrated on the right end of the histogram. Surprisingly, the distributions of complementary activities like \textit{Opening Facility Door} and \textit{Closing Facility Door} are not identical, which might be a source of bias. 

In \cref{tab:comparison}, we show the statistics on the MAMA dataset. MAMA consists of a total of 32726 video with 25837 videos in the train split, and 6889 videos in the test split. The length of clips in the dataset ranges from 1 sec to 5 sec.

\subsection{Difficulty And Diversity in MAMA Dataset}

\begin{figure*}[t]
\begin{tabular}{ccc}
\subcaptionbox{UCF101: BBox Area w.r.t Frame}{\includegraphics[width=0.3\linewidth]{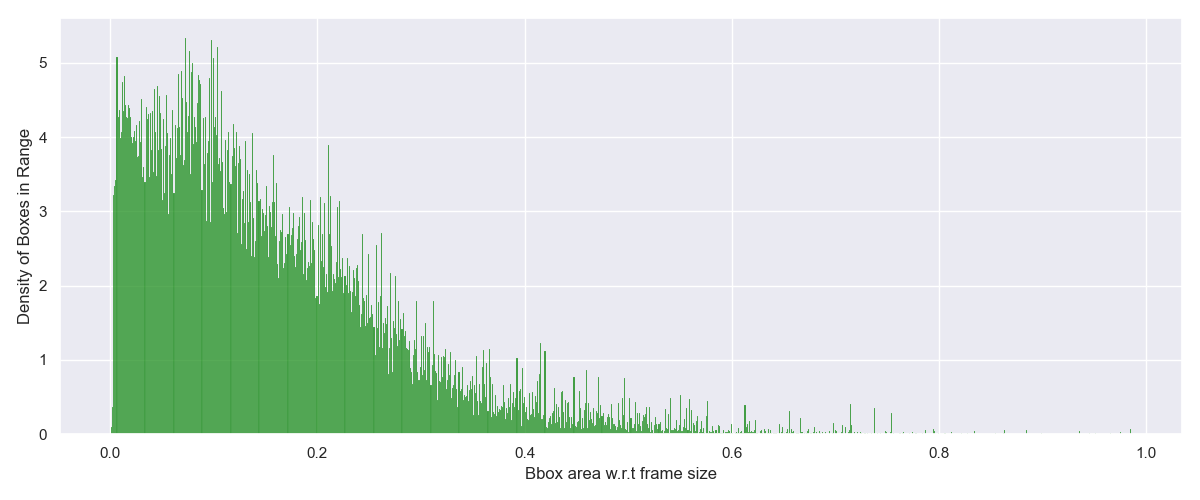}} &
\subcaptionbox{JHMDB: BBox Area w.r.t frame}{\includegraphics[width=0.3\linewidth]{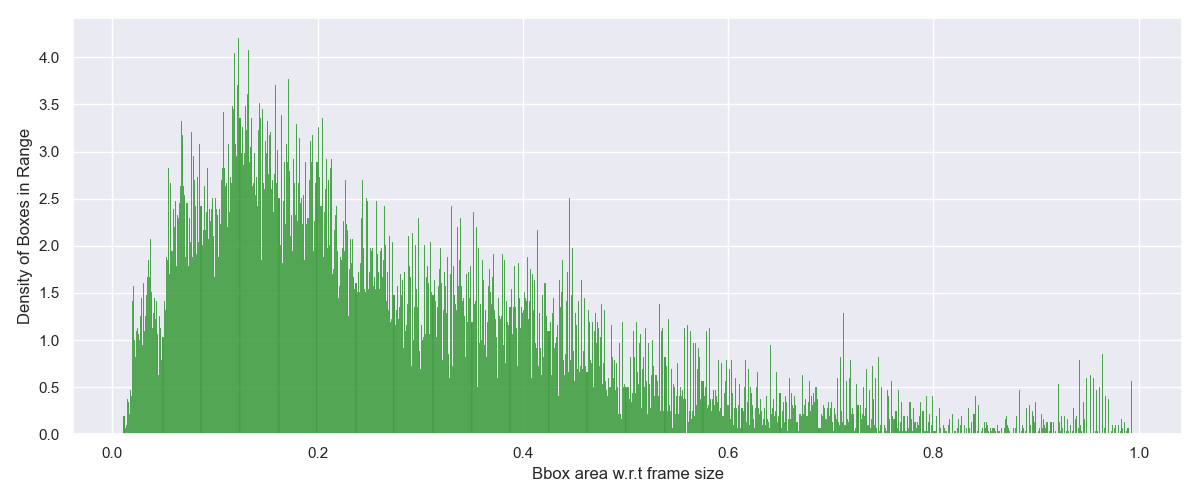}} &
\subcaptionbox{MAMA: BBox Area w.r.t frame}{\includegraphics[width=0.3\linewidth]{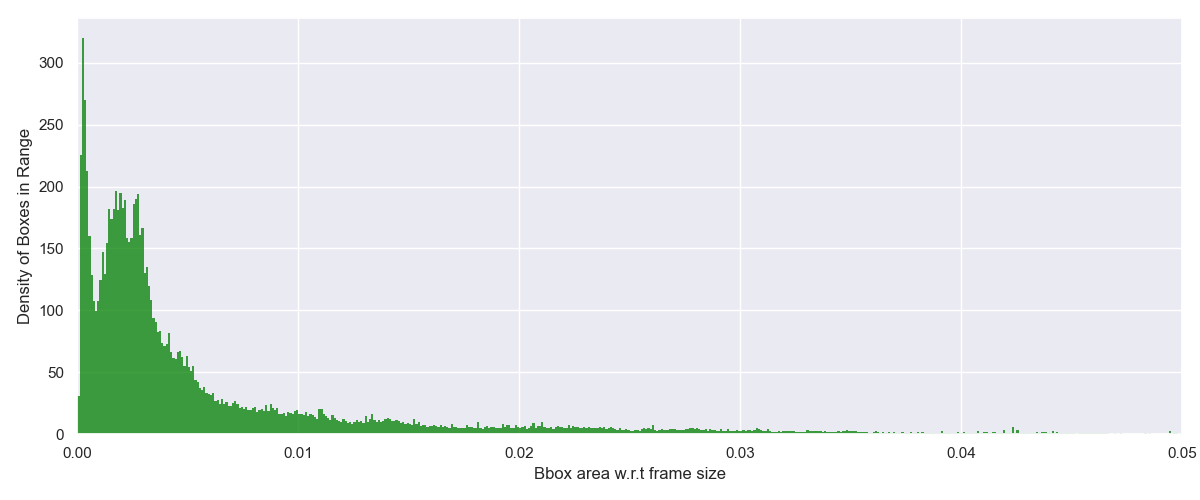}} \\
\subcaptionbox{UCF101: Aspect Ratio (h/w)}{\includegraphics[width=0.3\linewidth]{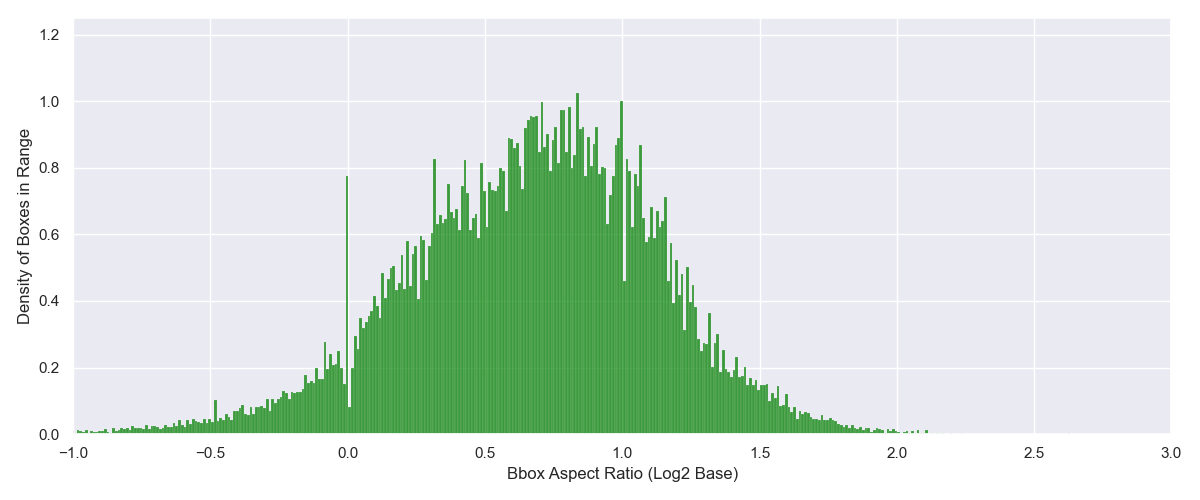}} &
\subcaptionbox{JHMDB: Aspect Ratio (h/w)}{\includegraphics[width=0.3\linewidth]{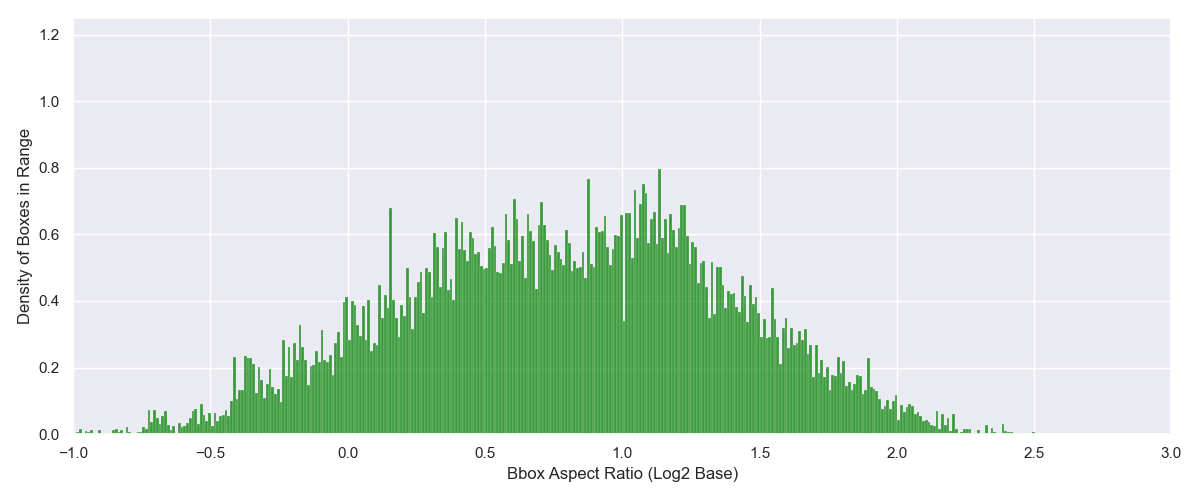}} &
\subcaptionbox{MAMA{$^{\wedge}$}: Aspect Ratio (h/w)}{\includegraphics[width=0.3\linewidth]{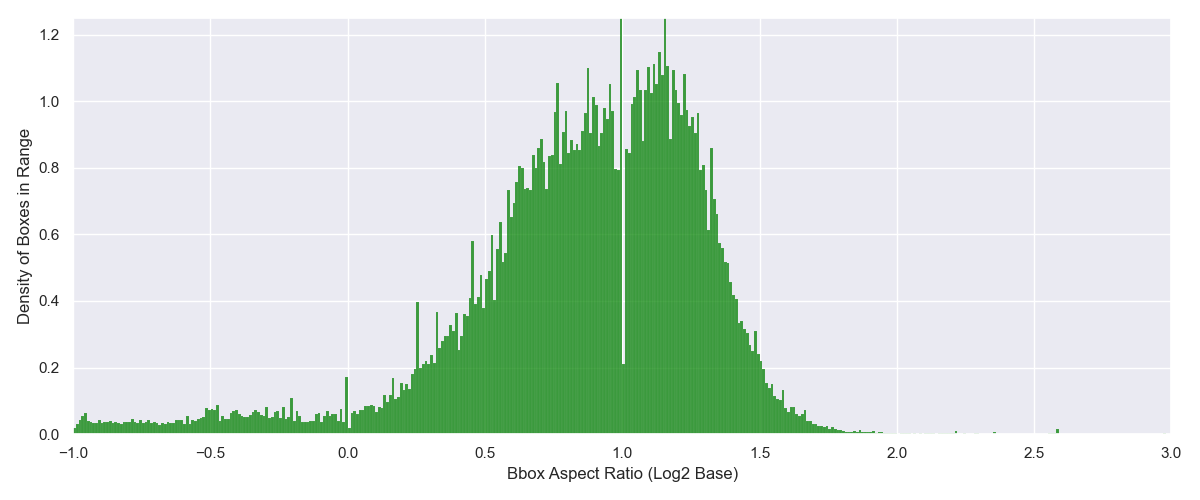}}
\end{tabular}
\vspace{-5pt}
 \caption{\small First Row: (X Axis) shows the possible areas of BBoxes w.r.t video frame in several action detection datasets. (Y Axis) shows the normalized density of bbox occurences. Notice how MAMA's density is concentrated on the left, signifying that most of the bboxes occupy lower resolution. Second Row: (X Axis): We plot the aspect ratios of the bboxes on a logarithmic scale. All the datasets follow  a gaussian distribution, but MAMA has lower variance. Hence MAMA dataset models low resolution conditions with higher probability.  $^{\wedge}$: X axis has been scaled to 0.05 due to lower bbox aspect ratios.}
\label{fig:density-comparisons}
\vspace{-5pt}
\end{figure*}



\begin{figure*}
    \centering
    \includegraphics[width=\linewidth]{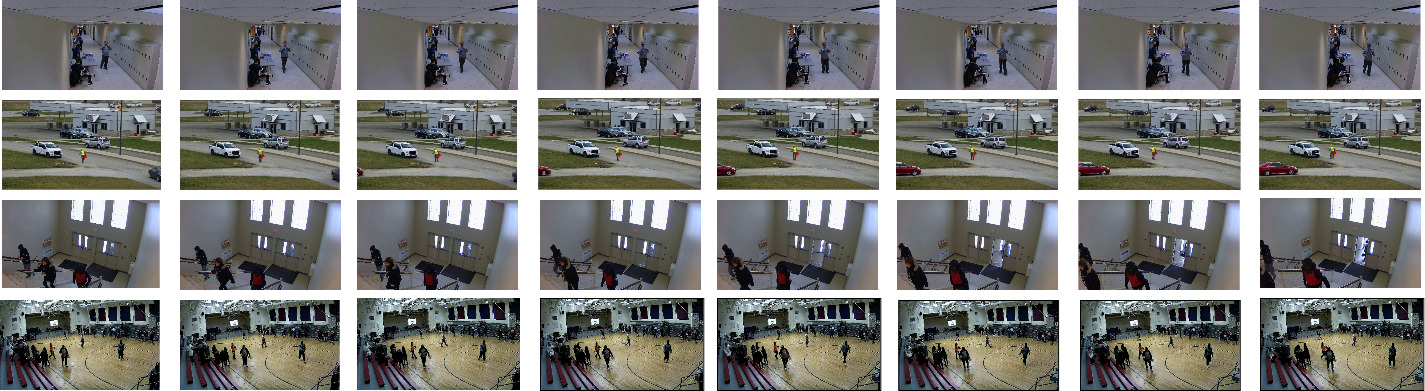}
    \caption{Samples from the proposed MAMA dataset are illustrated. The video frames span both indoor/outdoor scenes and  multiple actors. MAMA is realistic due to extremely fine temporal changes in actor trajectory. Best viewed on screen. }
    \label{fig:mama_samples}
\end{figure*}

\textbf{Indoor/Outdoor scenes:} MAMA dataset contains the trimmed clips from both VIRAT/MEVA datasets. Due to this, MAMA contains both indoor/outdoor scenes. Certain indoor scenes like \textit{Baseketball Court} are crowded, containing over 44 actors in a single frame. On the other hand, the outdoor scenes are generally relatively sparse. During the course of mapping the activities from VIRAT to MEVA during the dataset construction, we dropped certain labels like that of a \textit{Person Walking}. Our belief is that an activity is characterized only when a person \textit{interacts} with another object/person. Activities like walking just involves a single actor. So, our dataset will help models to learn to focus on \textit{activities of interest} rather than just focusing on 'all' temporal movements in a video. We present some of the samples from our dataset in Figure \ref{fig:mama_samples}.

\textbf{Bounding Box Areas and Aspect Ratios:} Figure \ref{fig:density-comparisons}\textcolor{red}{a}, shows the relative distribution of the bbox areas with respect to the frame area on a logarithmic scale. Most of the samples in the MAMA dataset are constrained to $<2\%$ of the frame area. Concurrently, Figure \ref{fig:density-comparisons}\textcolor{red}{b}, shows the density curve of the bounding box aspect ratios. While the bboxes generally model a nice gaussian distribution, we observe that the curve extends almost equally in both negative and positive ranges. This shows that MAMA dataset contains an almost equal no of wide and long objects.

\subsection{Evaluation Protocol}
We modify output channels of decoder in VideoCapsuleNet\cite{duarte2018videocapsulenet} to predict semantic segmentation volume for each of the 35 activities in the MAMA  dataset. Then, we run 3D connected components on each of the activity's volume to isolate the 3D tubes corresponding to an actor temporally.  Components with temporal length less than 4 frames and lesser than 20 pixel prediction per frame are dropped. Finally, we fit bboxes on per-pixel actor localization to obtain frame wise detections.

 Following the protocol as described in \cite{kalogeiton2017action}, we estimate the fMAP at 0.5 ioU threshold. Then, linking is done temporally to obtain the predicted tubes from the VideoCapsuleNet baseline. \cite{duarte2018videocapsulenet}. Finally, we estimate the spatio-temporal overlap between predicted and ground truth tubes using 3D ioU and report the vMAP. For classification accuracy, we report balanced accuracy over 35 classes of the MAMA dataset.

\begin{figure*}
    \centering
    \includegraphics[width=1\linewidth]{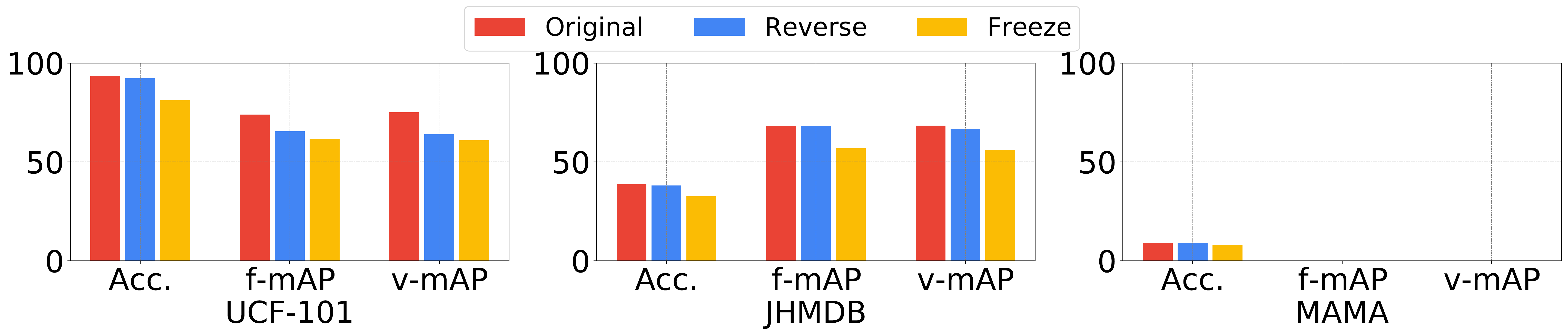}
    \caption{Overview of the effect of temporal aspect for different datasets. We show the accuracy, f-mAP and v-MAP (@0.5 IoU) for original, reversed and frozen samples across datasets.}
    \label{fig:all_dataset_compare}
\end{figure*}

\section{Analysing datasets}

Actions from videos takes into account the position of involved actors in space and time and their relation to the surrounding during the action. From a detection perspective, one key aspect is to understand how the actor changes with respect to their surrounding in temporal dimension. We look into how the temporal information can be influential for understanding different actions from different datasets.

\subsection{Importance of temporal aspect}
First we analyze to what degree does action understanding rely on the temporal aspect. Scene bias in videos is an issue which questions the need of temporal aspect in video understanding \cite{choi2019can}. Certain actions can be guessed based on the scene alone, which leads to a heavy scene bias during learning. This can limit the use of such dataset for generalizing action detection to real world videos. We evaluate how important the temporal information is for actions across various datasets available for action detection.

\subsection{Importance of temporal order}
Along with scene bias, we also look into the importance of the temporal order of the information in order to understand the activity. While actions are based on how the actor moves in space and time, the temporal order of the motion can be essential to understand how important the scenes from each dataset are to detect the actions. Getting this relation between temporal order and action understanding can give insight into developing methods for better action detection.


\section{Experiments and results}
We will briefly describe the statistics for datasets used in our work to compare with the new dataset. Then, we discuss recent approaches for activity detection, their limitations, and our proposed modifications as well.

\subsection{Datasets}

\noindent \textbf{UCF101-24: } The number of training and testing videos in UCF101 is 2.2k and 900 respectively. The videos are untrimmed and the resolution of videos is 320x240.

\noindent \textbf{JHMDB-21: } Total number of videos are approximately 900 which is divided into 2:1 for training and testing. Video resolution is same as UCF101-24, however, the videos are trimmed. 

\noindent \textbf{AVA: } dataset cotains long untrimmed videos distributed amongst 80 actions. The resolution of videos are 320x400. 

\noindent \textbf{MAMA: } The train test split is 25.8k and 6.9k respectively. The video resolution is 1920x1080.


For UCF101-24 and JHMDB-21, videos are resized to 224x224. For AVA, the frame size is 400x400. For MAMA dataset, the input resolution is 256x256. 
\subsection{Approaches}

\noindent \textbf{VCN: } \cite{duarte2018videocapsulenet} is an end-to-end 3d encoder decoder based approach. The features are extracted at multiple checkpoints and its fed into the capsule routing algorithm. Video is classified from the output of encoder. To generate the localization map, we upsample the encoder features also known as activations. The ground truth vector is multiplied by the activations to propagate the class information. However, this approach works only if the output is binary, meaning presence of a single action across the whole video. This approach is not suitable on AVA and MAMA dataset.

\noindent \textbf{STEP: } \cite{yang2019step} It progressively refines proposal using regression. Starting with rough estimate of proposals, the network, updates the proposals with each iteration. It also targets to extend the bounding boxes in temporal dimension.  The architecture has two branch: Global branch that works on spatio-temporal modeling of the whole input sequence and a local branch that applies bounding box regression on each frame. 


\noindent \textbf{VCN-MA:} We extend the original work
\cite{duarte2018videocapsulenet} from one channel to multiple channels. Since, we have 35 classes in our dataset, we have in total 36 (35 activity classes + 1 background channel) channels. We also extended the spread loss such that it's applicable for multiple dimensions. \cite{hinton2018matrix} promotes the activation of the target class to be far away from the other classes by a margin factor. It does not 'force' the activations of the wrong classes to be far apart. We implement a multi dimensional variant, where all logits corresponding to  multiple ground truth action classes are far away with significant margin, and retain this non-separation behaviour in the rest.


\subsection{Evaluation metrics}

We show performance on three metrics classification accuracy, f-mAP and v-mAP. Given IoU value $p$, f-mAP metric provides information about how many predicted frames have atleast spatial overlap of $p$ from the ground truth. Similarly, v-mAP provides information about spatio-temporal overlap for different values of IoU.

To further analyze across different video datasets, it's important to ponder over how much scene bias is in the dataset, and, how important is the temporal ordering of frames. Thus, we perform two sets of experiments: 1) Freeze: We record the performance by taking the center frame and repeating it for the total number of frames present in the original video to analyze scene bias, 2) Reversal: We reverse the order of frames along temporal dimension. Then, we measure the absolute $(\sigma - \sigma^{'})$ and relative $((\sigma - \sigma^{'})/ \sigma)$ drop from the original scores for accuracy, f-mAP and v-mAP. $\sigma$ for f-mAP calculation is shown in equation \ref{eqn:fmap_cal}.


\begin{equation}
\footnotesize
    \sigma_{0.5} = \frac{\sum_{i=1}^{n} (fmap>0.5)}{n}
    \label{eqn:fmap_cal}
\end{equation}
where $\sigma_{0.5}$ means it assigns 1 if IoU value is greater than 0.5 and $n$ denotes the number of frames. $fmap$ is defined in equation \ref{eqn:fmap_calculation}. $A_{P}$ and $A_{GT}$ means predicted area and ground truth area.
\begin{equation}
\footnotesize
    fmap = \frac{|A_{P}\cap A_{GT}|}{|A_{P}\cup  A_{GT}|} 
    \label{eqn:fmap_calculation}
\end{equation}

\subsection{Results}
After training VCN-MA for 100 epochs, we report accuracy, fmap and vmap scores in Table \ref{tab:compare_all}. Additionally, we evaluate classification performance on several datasets in Table \ref{tab:compare_classifiers}.

Further, we investigate scene bias and importance of temporal ordering.  Firstly, for scene bias, we see the most absolute and relative drop in performance for f-mAP@0.5 for AVA dataset. For v-mAP@0.5, UCF101-24 has the most absolute and relative drop. Since, the network backbone is same, we reverse the order of frames. We flip the clip along the temporal dimension and then compare the performance with the original input. From Tables \ref{tab:freeze_drops} and \ref{tab:reversal_drops}, we can see that the relative drop in f-mAP@0.5 is most for UCF101-24. In case of JHMDB-21, there's no drop in performance for f-mAP. We see the similar trend for AVA dataset as well. The performance drop is 1.2\% but relative drop in performance is high. 

\begin{figure}[t]
  \centering
\includegraphics[width=\linewidth]{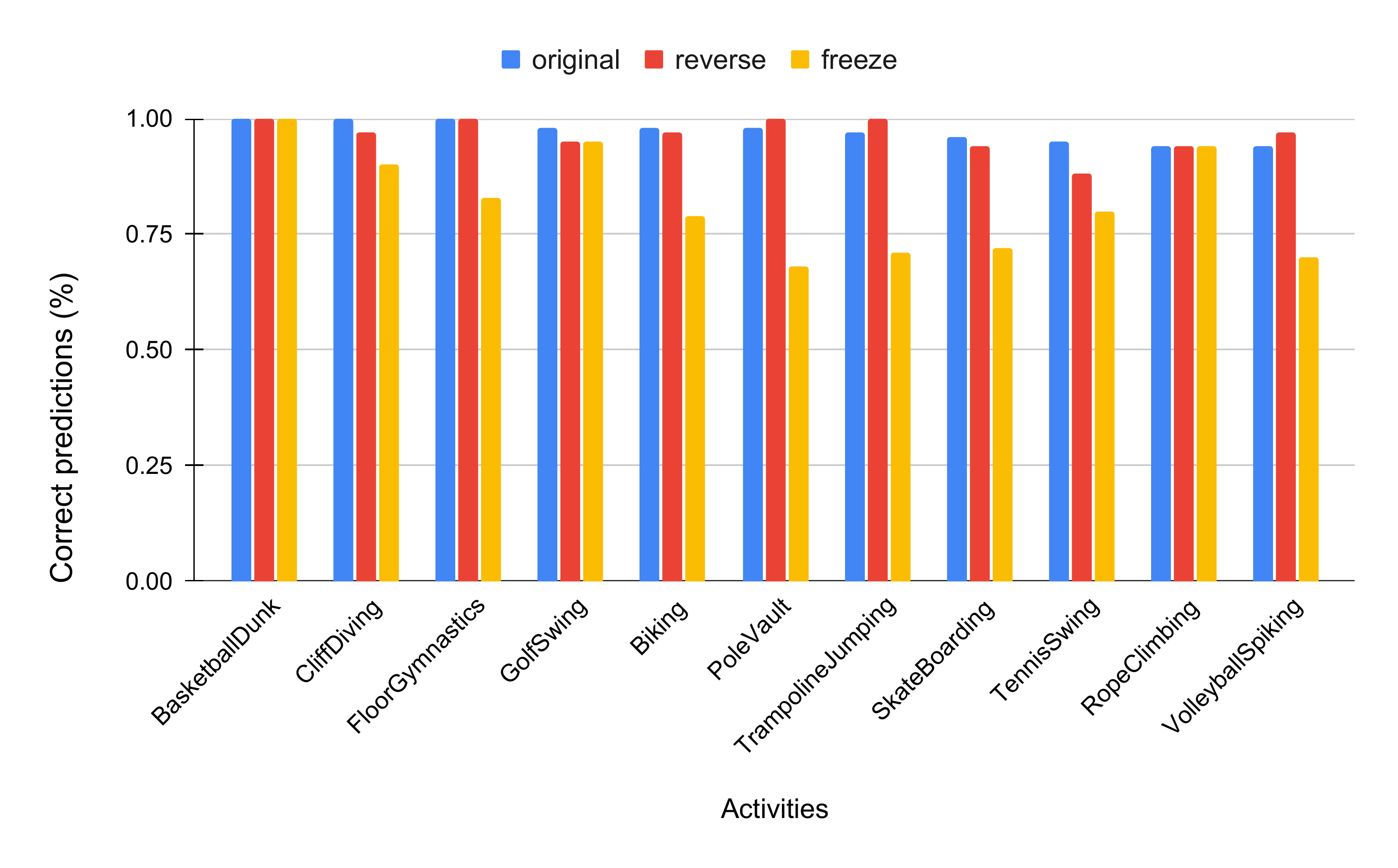}
  \caption{UCF101-24 - Frame order class accuracy. }
  \label{fig:ucf_order}
\end{figure}

\begin{figure}[t]
  \centering
\includegraphics[width=\linewidth]{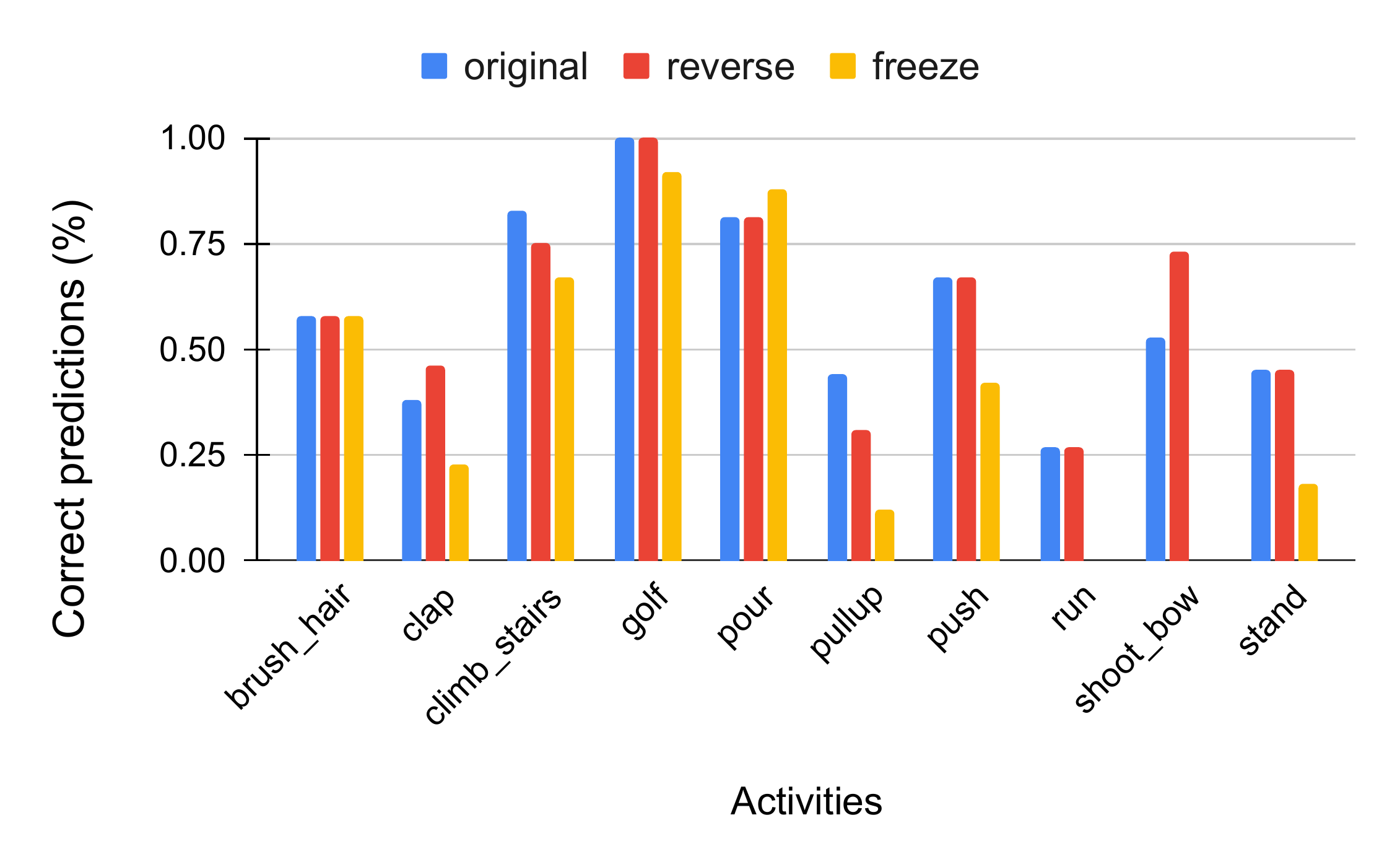}
  \caption{JHMDB-21 - Frame order class accuracy. }
  \label{fig:jhmdb_order}
\end{figure}

Next, we analyze the class accuracy under different types of evaluations. We look into top 10 classes with highest score in normal evaluation. In UCF101-24 (Fig. \ref{fig:ucf_order}), we see the drop in scores for frames freeze case, when there's a long interaction between actor and object (e.g. \texttt{polevault}, \texttt{biking}, \texttt{trampoline\_jump}, and \texttt{skateboarding}). In JHMDB-21(Fig. \ref{fig:jhmdb_order}), we see performance drop specifically for activities that involves fast motion such as \texttt{running} and \texttt{shooting bow} in case of freezing the frames. In AVA dataset, \texttt{walking}, \texttt{driving} and\texttt{swimming} has almost zero accuracy in case of freezing the frame. If there's little or no motion, then the classification accuracy is comparable with normal evaluation protocol for example \texttt{sit}, \texttt{lie}, \texttt{stand}, and \texttt{watch}. Figure \ref{fig:all_dataset_compare} illustrates the relative comparison of freezing and reversing the input frames on several datasets.  

\begin{figure}[t]
  \centering
\includegraphics[width=\linewidth]{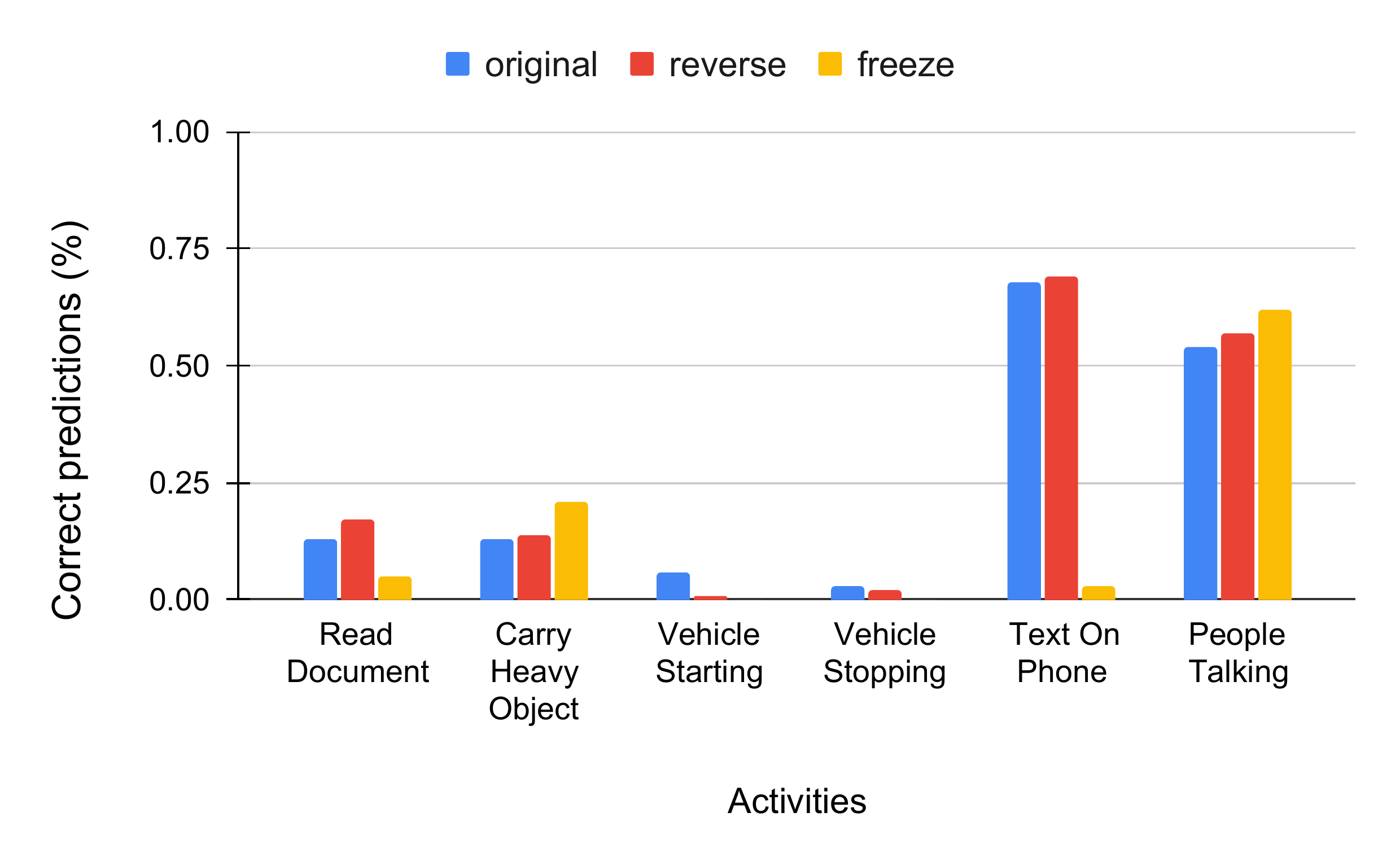}
  \caption{MAMA - Frame order class accuracy. }
  \label{fig:mama_order}
\end{figure}


\begin{table}
    \centering
    \begin{tabular}{c| c|c|c|c}
    
    Dataset & UCF101-24 &JHMDB-21 &  AVA\textsuperscript{\textdagger} &  MAMA{$^{\wedge}$}  \\
    \hline
    Acc.  &  83.2 &  34.3&  21.2 &  4.1\\
  \end{tabular}
  \caption{Performance comparison for classifiers trained on different datasets using I3D backbone. {\textdagger}: AVA - (STEP - I3D). $^{\wedge}$: MAMA - (VCN-MA - I3D). }
  \label{tab:compare_classifiers}
\end{table}

\begin{table}
    \centering
    \begin{tabular}{c c ccc}
    Dataset &  Order & Acc & f-mAP & v-mAP \\
    \midrule 
    
    \multirow{3}{*}{UCF101-24} & Original& 93.4 &74.0 & 75.1\\
     &Freeze & 81.2 & 61.7 & 60.9 \\
     & Reverse & 92.2 & 65.5 & 63.9 \\
    \midrule
    \multirow{3}{*}{JHMDB-21} & Original& 38.8 & 68.2 & 68.4\\
     &Freeze & 32.6 & 56.9 &56.2 \\
     & Reverse & 38.1 & 68.1 & 66.7\\
    \midrule
    \multirow{3}{*}{AVA\textsuperscript{\textdagger}} & Original& 21.2 &20.1 & -\\
     &Freeze & 7.7 & 7.4 & - \\
     & Reverse & 20.8 & 18.9 & -\\
     \midrule
     \multirow{3}{*}{MAMA{$^{\wedge}$}} & Original& 9.2 & 0.4& 0.1 \\
     &Freeze &  8.1 & 0.1& 0.0\\
     & Reverse & 9.2 & 0.1& 0.0\\
  \end{tabular}
  \caption{Performance comparison for different types of frame ordering at f-mAP@0.5 and v-mAP@0.5. \textsuperscript{\textdagger}: AVA results are shown for illustration purposes only. $^{\wedge}$: MAMA is a challenging dataset, and our baseline models yields low accuracy.}
  \label{tab:compare_all}
\end{table}

\begin{table}
    \centering
    \begin{tabular}{c c ccc}
    Dataset &  Order & Acc & f-mAP & v-mAP \\
    \midrule 
    \multirow{4}{*}{Absolute} & UCF101-24& 12.2 & 12.3 & 14.2\\
     &JHMDB-21 & 6.2 & 11.3 & 12.2 \\
     & AVA\textsuperscript{\textdagger} & \textbf{13.5} & \textbf{12.7} & - \\
     & MAMA & 1.1 & 0.3 & 0.07\\
    \midrule
    \multirow{4}{*}{Relative} & UCF101-24& 13.1& 16.6 & 18.9\\
     &JHMDB-21 & 16.0 & 16.5 & 17.8 \\
     & AVA\textsuperscript{\textdagger} & \textbf{63.7} & 63.2 & - \\
     & MAMA & 12.0 & \textbf{75} & - \\
  \end{tabular}
  \caption{Absolute and relative drop in performance for scene bias condition at f-mAP@0.5 and v-mAP@0.5.\textsuperscript{\textdagger}: AVA results are shown for illustration purposes only.}
  \label{tab:freeze_drops}
\end{table}

\begin{table}
    \centering
    \begin{tabular}{c c ccc}
    Dataset &  Order & Acc & f-mAP & v-mAP \\
    \midrule 
    \multirow{4}{*}{Absolute} & UCF101-24& \textbf{1.2} & \textbf{8.7} & 11.2\\
     &JHMDB-21 & 0.7 & 0.1 & 1.7\\
     & AVA\textsuperscript{\textdagger} & 0.4 & 1.2 & - \\
     & MAMA & 0.0 & 0.3 & 0.08\\
    \midrule
    \multirow{4}{*}{Relative} & UCF101-24& 1.3 & 11.8 & 14.9\\
     &JHMDB-21 & \textbf{1.8} & 0.0 & 2.4\\
     & AVA\textsuperscript{\textdagger} & 0.0 & 6.0 & - \\
     & MAMA & 0.0 & \textbf{75} & -\\
  \end{tabular}
  \caption{Absolute and relative drop in performance for temporal reverse condition at f-mAP@0.5 and v-mAP@0.5. \textsuperscript{\textdagger}: AVA results are shown for illustration purposes only. }
  \label{tab:reversal_drops}
\end{table}

\subsection{Challenges and Discussions}


In this section, we list out several key challenges MAMA dataset list out that corroborates to real-world scenarios: 

\noindent \textbf{Video resolution:} Most of the video datasets are low resolution, thus, action detection approaches are designed with input size of 224 or 256 resolution.  Resizing a 1920x1080 video to 224x224 leads to information loss which makes proposed approaches to work efficiently on this dataset. \\
\textbf{Class imbalance}: Diving deeper into real-world scenario, there's always a huge class imbalance in activities. Some activities are more frequent than others. For example, \texttt{talking} and \texttt{texting} are activities have a higher frequency than activities such as \texttt{close\_trunk}, \texttt{people\_embrace}, and, \texttt{close\_facility\_door}.  \\
\textbf{Instance separation:} Existing datasets have been collected under controlled number of instances. However, number of instances vary by a good margin between an indoor scene of a cafeteria vs an outdoor scene of a parking lot. A lot of actors are concentrated in a smaller area vs less actors spread over a large area.


\section{Related work}
\cite{barker1955midwest} discusses the hierarchical organizations of activities, specifically breaking down an activity into varying types of behaviours exhibited by an actor. Motivated by this, datasets like AVA \cite{gu2018ava}, classify actions as 'atomic' and try to capture subtle changes in actor motion by annotating at 1 sec intervals. On the other hand, datasets like UCF101\cite{soomro2012ucf101}, JHMDB\cite{jhuang2013towards_jhmdb} perform annotations at finer spatio temporal scales. However, such datasets do not capture the subtle atomic actions. Simlarly, a lot of relevant work \cite{tartaglione2021end} \cite{choi2019can} has been done on detecting and removing bias by training networks on masked actor regions. However, understanding  the sources of bias in the context of action \textit{detection} is a relatively unexplored problem.  

\section{Conclusion}
We have analyzed existing video datasets and shown the importance of temporal aspects for action understanding. Specifically, we have demonstrated how a detection model behaves when the temporal information is frozen or reversed. A reliable metric to compare datasets could be to measure the relative drops of models across datasets on several such properties. Furthermore, to contribute to  the ongoing research in action detection, we have presented a new spatio-temporal dataset titled MAMA.

{\small
\bibliographystyle{ieee_fullname}
\bibliography{egbib}

\begin{thebibliography}{10}\itemsep=-1pt

\bibitem{barker1955midwest}
Roger~G Barker and Herbert~F Wright.
\newblock Midwest and its children: The psychological ecology of an american
  town.
\newblock 1955.

\bibitem{caba2015activitynet}
Fabian Caba~Heilbron, Victor Escorcia, Bernard Ghanem, and Juan Carlos~Niebles.
\newblock Activitynet: A large-scale video benchmark for human activity
  understanding.
\newblock In {\em Proceedings of the ieee conference on computer vision and
  pattern recognition}, pages 961--970, 2015.

\bibitem{carreira2017quo_i3d}
Joao Carreira and Andrew Zisserman.
\newblock Quo vadis, action recognition? a new model and the kinetics dataset.
\newblock In {\em proceedings of the IEEE Conference on Computer Vision and
  Pattern Recognition}, pages 6299--6308, 2017.

\bibitem{choi2019can}
Jinwoo Choi, Chen Gao, Joseph~CE Messou, and Jia-Bin Huang.
\newblock Why can't i dance in the mall? learning to mitigate scene bias in
  action recognition.
\newblock {\em Advances in Neural Information Processing Systems}, 32, 2019.

\bibitem{corona2021meva}
Kellie Corona, Katie Osterdahl, Roderic Collins, and Anthony Hoogs.
\newblock Meva: A large-scale multiview, multimodal video dataset for activity
  detection.
\newblock In {\em Proceedings of the IEEE/CVF Winter Conference on Applications
  of Computer Vision}, pages 1060--1068, 2021.

\bibitem{dave2022gabriellav2}
Ishan Dave, Zacchaeus Scheffer, Akash Kumar, Sarah Shiraz, Yogesh~Singh Rawat,
  and Mubarak Shah.
\newblock Gabriellav2: Towards better generalization in surveillance videos for
  action detection.
\newblock In {\em Proceedings of the IEEE/CVF Winter Conference on Applications
  of Computer Vision}, pages 122--132, 2022.

\bibitem{demir2021tinyvirat}
Ugur Demir, Yogesh~S Rawat, and Mubarak Shah.
\newblock Tinyvirat: Low-resolution video action recognition.
\newblock In {\em 2020 25th International Conference on Pattern Recognition
  (ICPR)}, pages 7387--7394. IEEE, 2021.

\bibitem{duarte2018videocapsulenet}
Kevin Duarte, Yogesh Rawat, and Mubarak Shah.
\newblock Videocapsulenet: A simplified network for action detection.
\newblock In {\em Advances in Neural Information Processing Systems}, pages
  7610--7619, 2018.

\bibitem{el2018real}
Alaaeldin El-Nouby and Graham~W Taylor.
\newblock Real-time end-to-end action detection with two-stream networks.
\newblock {\em arXiv preprint arXiv:1802.08362}, 2018.

\bibitem{fan2021multiscale}
Haoqi Fan, Bo Xiong, Karttikeya Mangalam, Yanghao Li, Zhicheng Yan, Jitendra
  Malik, and Christoph Feichtenhofer.
\newblock Multiscale vision transformers.
\newblock In {\em Proceedings of the IEEE/CVF International Conference on
  Computer Vision}, pages 6824--6835, 2021.

\bibitem{feichtenhofer2019slowfast}
Christoph Feichtenhofer, Haoqi Fan, Jitendra Malik, and Kaiming He.
\newblock Slowfast networks for video recognition.
\newblock In {\em Proceedings of the IEEE/CVF international conference on
  computer vision}, pages 6202--6211, 2019.

\bibitem{gu2018ava}
Chunhui Gu, Chen Sun, David~A Ross, Carl Vondrick, Caroline Pantofaru, Yeqing
  Li, Sudheendra Vijayanarasimhan, George Toderici, Susanna Ricco, Rahul
  Sukthankar, et~al.
\newblock Ava: A video dataset of spatio-temporally localized atomic visual
  actions.
\newblock In {\em Proceedings of the IEEE Conference on Computer Vision and
  Pattern Recognition}, pages 6047--6056, 2018.

\bibitem{hara2018can_r3d}
Kensho Hara, Hirokatsu Kataoka, and Yutaka Satoh.
\newblock Can spatiotemporal 3d cnns retrace the history of 2d cnns and
  imagenet?
\newblock In {\em Proceedings of the IEEE conference on Computer Vision and
  Pattern Recognition}, pages 6546--6555, 2018.

\bibitem{hinton2018matrix}
Geoffrey~E Hinton, Sara Sabour, and Nicholas Frosst.
\newblock Matrix capsules with em routing.
\newblock In {\em International conference on learning representations}, 2018.

\bibitem{hou2017tcnn}
Rui Hou, Chen Chen, and Mubarak Shah.
\newblock Tube convolutional neural network (t-cnn) for action detection in
  videos.
\newblock In {\em IEEE International Conference on Computer Vision}, 2017.

\bibitem{idrees2017thumos}
Haroon Idrees, Amir~R Zamir, Yu-Gang Jiang, Alex Gorban, Ivan Laptev, Rahul
  Sukthankar, and Mubarak Shah.
\newblock The thumos challenge on action recognition for videos “in the
  wild”.
\newblock {\em Computer Vision and Image Understanding}, 155:1--23, 2017.

\bibitem{jhuang2013towards_jhmdb}
Hueihan Jhuang, Juergen Gall, Silvia Zuffi, Cordelia Schmid, and Michael~J
  Black.
\newblock Towards understanding action recognition.
\newblock In {\em Proceedings of the IEEE international conference on computer
  vision}, pages 3192--3199, 2013.

\bibitem{kalogeiton2017action}
Vicky Kalogeiton, Philippe Weinzaepfel, Vittorio Ferrari, and Cordelia Schmid.
\newblock Action tubelet detector for spatio-temporal action localization.
\newblock In {\em Proceedings of the IEEE International Conference on Computer
  Vision}, pages 4405--4413, 2017.

\bibitem{karpathy2014large}
Andrej Karpathy, George Toderici, Sanketh Shetty, Thomas Leung, Rahul
  Sukthankar, and Li Fei-Fei.
\newblock Large-scale video classification with convolutional neural networks.
\newblock In {\em Proceedings of the IEEE conference on Computer Vision and
  Pattern Recognition}, pages 1725--1732, 2014.

\bibitem{kuehne2011hmdb}
Hildegard Kuehne, Hueihan Jhuang, Est{\'\i}baliz Garrote, Tomaso Poggio, and
  Thomas Serre.
\newblock Hmdb: a large video database for human motion recognition.
\newblock In {\em 2011 International conference on computer vision}, pages
  2556--2563. IEEE, 2011.

\bibitem{kumar2022end}
Akash Kumar and Yogesh~Singh Rawat.
\newblock End-to-end semi-supervised learning for video action detection.
\newblock {\em IEEE conference on computer vision and pattern recognition},
  2022.

\bibitem{li2018recurrent}
Dong Li, Zhaofan Qiu, Qi Dai, Ting Yao, and Tao Mei.
\newblock Recurrent tubelet proposal and recognition networks for action
  detection.
\newblock In {\em Proceedings of the European conference on computer vision
  (ECCV)}, pages 303--318, 2018.

\bibitem{mcintosh2020visual}
Bruce McIntosh, Kevin Duarte, Yogesh~S Rawat, and Mubarak Shah.
\newblock Visual-textual capsule routing for text-based video segmentation.
\newblock In {\em Proceedings of the IEEE/CVF Conference on Computer Vision and
  Pattern Recognition}, pages 9942--9951, 2020.

\bibitem{monfort2019moments}
Mathew Monfort, Alex Andonian, Bolei Zhou, Kandan Ramakrishnan, Sarah~Adel
  Bargal, Tom Yan, Lisa Brown, Quanfu Fan, Dan Gutfreund, Carl Vondrick, et~al.
\newblock Moments in time dataset: one million videos for event understanding.
\newblock {\em IEEE transactions on pattern analysis and machine intelligence},
  42(2):502--508, 2019.

\bibitem{oh2011large_virat}
Sangmin Oh, Anthony Hoogs, Amitha Perera, Naresh Cuntoor, Chia-Chih Chen,
  Jong~Taek Lee, Saurajit Mukherjee, JK Aggarwal, Hyungtae Lee, Larry Davis,
  et~al.
\newblock A large-scale benchmark dataset for event recognition in surveillance
  video.
\newblock In {\em CVPR 2011}, pages 3153--3160. IEEE, 2011.

\bibitem{rana2021we}
Aayush~J Rana and Yogesh~S Rawat.
\newblock We don't need thousand proposals: Single shot actor-action detection
  in videos.
\newblock In {\em Proceedings of the IEEE/CVF Winter Conference on Applications
  of Computer Vision}, pages 2960--2969, 2021.

\bibitem{rizve2021gabriella}
Mamshad~Nayeem Rizve, Ugur Demir, Praveen Tirupattur, Aayush~Jung Rana, Kevin
  Duarte, Ishan~R Dave, Yogesh~S Rawat, and Mubarak Shah.
\newblock Gabriella: An online system for real-time activity detection in
  untrimmed security videos.
\newblock In {\em 2020 25th International Conference on Pattern Recognition
  (ICPR)}, pages 4237--4244. IEEE, 2021.

\bibitem{rodriguez2008action_ucfsports}
Mikel~D Rodriguez, Javed Ahmed, and Mubarak Shah.
\newblock Action mach a spatio-temporal maximum average correlation height
  filter for action recognition.
\newblock In {\em 2008 IEEE conference on computer vision and pattern
  recognition}, pages 1--8. IEEE, 2008.

\bibitem{sigurdsson2016hollywood}
Gunnar~A Sigurdsson, G{\"u}l Varol, Xiaolong Wang, Ali Farhadi, Ivan Laptev,
  and Abhinav Gupta.
\newblock Hollywood in homes: Crowdsourcing data collection for activity
  understanding.
\newblock In {\em European Conference on Computer Vision}, pages 510--526.
  Springer, 2016.

\bibitem{soomro2014sports}
Khurram Soomro and Amir~R Zamir.
\newblock Action recognition in realistic sports videos.
\newblock In {\em Computer Vision in Sports}, pages 181--208. Springer, 2014.

\bibitem{soomro2012ucf101}
Khurram Soomro, Amir~Roshan Zamir, and Mubarak Shah.
\newblock Ucf101: A dataset of 101 human actions classes from videos in the
  wild.
\newblock {\em arXiv preprint arXiv:1212.0402}, 2012.

\bibitem{tartaglione2021end}
Enzo Tartaglione, Carlo~Alberto Barbano, and Marco Grangetto.
\newblock End: Entangling and disentangling deep representations for bias
  correction.
\newblock In {\em Proceedings of the IEEE/CVF Conference on Computer Vision and
  Pattern Recognition}, pages 13508--13517, 2021.

\bibitem{tirupattur2021tinyaction}
Praveen Tirupattur, Aayush~J Rana, Tushar Sangam, Shruti Vyas, Yogesh~S Rawat,
  and Mubarak Shah.
\newblock Tinyaction challenge: Recognizing real-world low-resolution
  activities in videos.
\newblock {\em arXiv preprint arXiv:2107.11494}, 2021.

\bibitem{tran2015c3dfb}
Du Tran, Lubomir Bourdev, Rob Fergus, Lorenzo Torresani, and Manohar Paluri.
\newblock Learning spatiotemporal features with 3d convolutional networks.
\newblock In {\em Computer Vision (ICCV), 2015 IEEE International Conference
  on}, pages 4489--4497. IEEE, 2015.

\bibitem{tran2018closer_r2p1d}
Du Tran, Heng Wang, Lorenzo Torresani, Jamie Ray, Yann LeCun, and Manohar
  Paluri.
\newblock A closer look at spatiotemporal convolutions for action recognition.
\newblock In {\em Proceedings of the IEEE conference on Computer Vision and
  Pattern Recognition}, pages 6450--6459, 2018.

\bibitem{vyas2020multi}
Shruti Vyas, Yogesh~S Rawat, and Mubarak Shah.
\newblock Multi-view action recognition using cross-view video prediction.
\newblock In {\em European Conference on Computer Vision}, pages 427--444.
  Springer, 2020.

\bibitem{wang2018nonlocal}
Xiaolong Wang, Ross Girshick, Abhinav Gupta, and Kaiming He.
\newblock Non-local neural networks.
\newblock In {\em Proceedings of the IEEE conference on computer vision and
  pattern recognition}, pages 7794--7803, 2018.

\bibitem{weinzaepfel2015learning}
Philippe Weinzaepfel, Zaid Harchaoui, and Cordelia Schmid.
\newblock Learning to track for spatio-temporal action localization.
\newblock In {\em Proceedings of the IEEE international conference on computer
  vision}, pages 3164--3172, 2015.

\bibitem{weinzaepfel2016human}
Philippe Weinzaepfel, Xavier Martin, and Cordelia Schmid.
\newblock Human action localization with sparse spatial supervision.
\newblock {\em arXiv preprint arXiv:1605.05197}, 2016.

\bibitem{xie2018rethinking}
Saining Xie, Chen Sun, Jonathan Huang, Zhuowen Tu, and Kevin Murphy.
\newblock Rethinking spatiotemporal feature learning: Speed-accuracy trade-offs
  in video classification.
\newblock In {\em Proceedings of the European conference on computer vision
  (ECCV)}, pages 305--321, 2018.

\bibitem{xu2015can}
Chenliang Xu, Shao-Hang Hsieh, Caiming Xiong, and Jason~J Corso.
\newblock Can humans fly? action understanding with multiple classes of actors.
\newblock In {\em Proceedings of the IEEE Conference on Computer Vision and
  Pattern Recognition}, pages 2264--2273, 2015.

\bibitem{yang2019step}
Xitong Yang, Xiaodong Yang, Ming-Yu Liu, Fanyi Xiao, Larry~S Davis, and Jan
  Kautz.
\newblock Step: Spatio-temporal progressive learning for video action
  detection.
\newblock In {\em Proceedings of the IEEE Conference on Computer Vision and
  Pattern Recognition}, pages 264--272, 2019.

\bibitem{yeung2018every}
Serena Yeung, Olga Russakovsky, Ning Jin, Mykhaylo Andriluka, Greg Mori, and Li
  Fei-Fei.
\newblock Every moment counts: Dense detailed labeling of actions in complex
  videos.
\newblock {\em International Journal of Computer Vision}, 126(2):375--389,
  2018.

\bibitem{yu2022argus++}
Lijun Yu, Yijun Qian, Wenhe Liu, and Alexander~G Hauptmann.
\newblock Argus++: Robust real-time activity detection for unconstrained video
  streams with overlapping cube proposals.
\newblock In {\em Proceedings of the IEEE/CVF Winter Conference on Applications
  of Computer Vision}, pages 112--121, 2022.

\end{thebibliography}
}

\end{document}